%% file: main.tex
\documentclass[letterpaper,10pt,conference]{ieeeconf}

\IEEEoverridecommandlockouts
\input{ICRA_Submission/0-macro}

\title{\method: Learning Whole-Body World Action Models with Scalable Motion Priors}

\author{
\begin{tabular}{c}
\textbf{Bowei Zhang}$^{1,2}$ \quad
\textbf{Qiyao Zhang}$^{2,3}$ \quad
\textbf{Shuanghao Bai}$^{2}$ \quad
\textbf{Xinhua Wang}$^{2}$ \quad
\textbf{Meng Li}$^{2,\dagger}$
\\[2pt]
\textbf{Yilei Wang}$^{4}$ \quad
\textbf{Leiwang Zhang}$^{4}$ \quad
\textbf{Jian Tang}$^{2}$ \quad
\textbf{Lu Zhou}$^{1,\ddagger}$ \quad
\textbf{Lei Sun}$^{1,\ddagger}$ \quad
\textbf{Zhengping Che}$^{2,\ddagger}$
\\[5pt]
{\small
$^{1}$Nankai University
\qquad
$^{2}$Beijing Innovation Center of Humanoid Robotics
}
\\[-1pt]
{\small
$^{3}$Beijing Institute of Technology
\qquad
$^{4}$Tsinghua University
}
\\[2pt]
{\small
$^{\dagger}$Project Lead
\qquad
$^{\ddagger}$Corresponding Authors
}
\end{tabular}
}
\usepackage{xcolor}
\begin{document}
\bstctlcite{IEEEexample:BSTcontrol}

\maketitle
\thispagestyle{empty}
\pagestyle{empty}

\begin{strip}
\vspace{-2.5\baselineskip} 
\centering
\includegraphics[width=\textwidth]{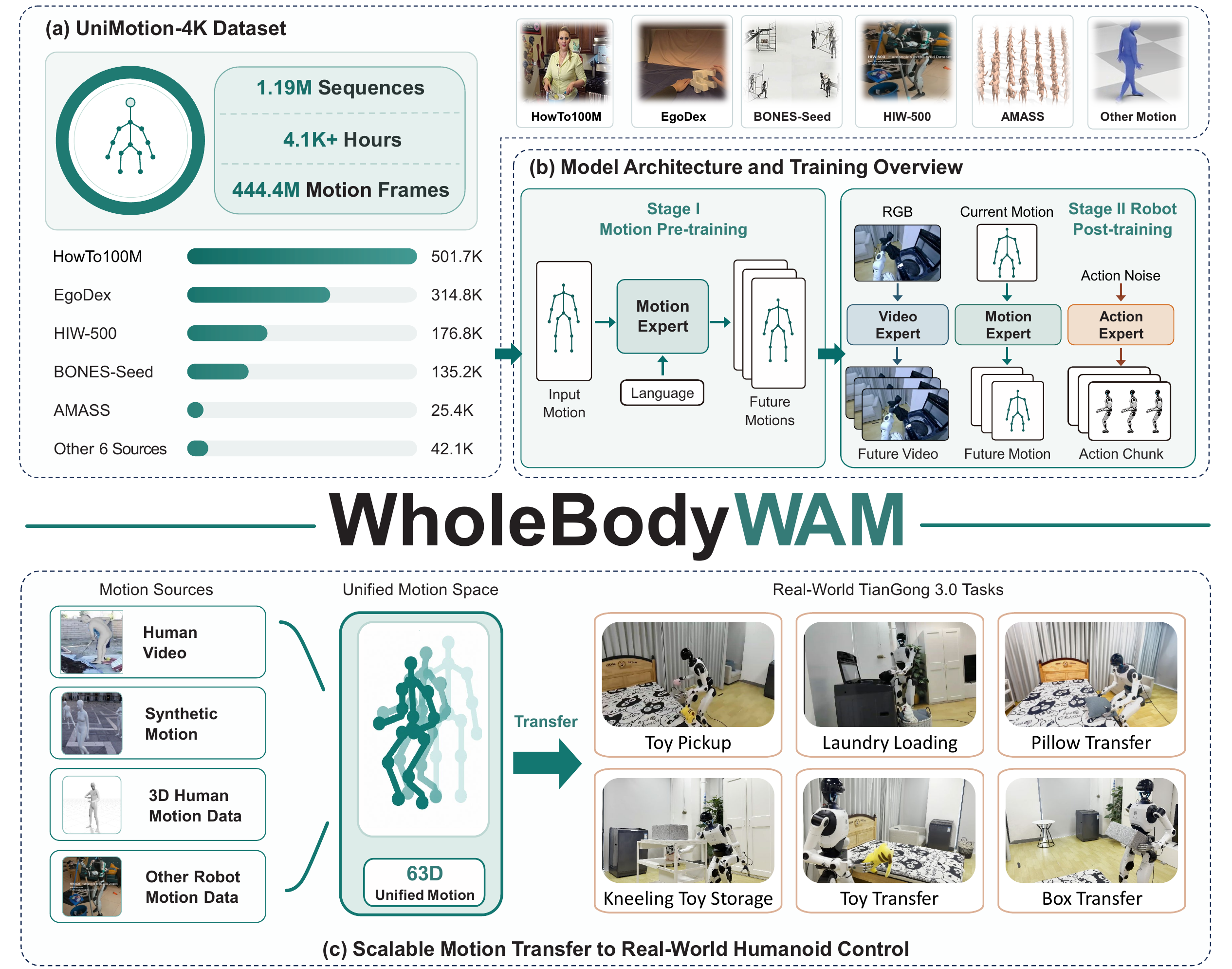}
\vskip -0.07in
\captionof{figure}{\textbf{Overview of \method.}
\method learns a scalable predictive whole-body motion prior from heterogeneous human and humanoid motion and transfers it to real-world humanoid control.
(a) We construct UniMotion-4K, comprising 1.19M motion sequences, 444.4M frames, and over 4.1K hours from diverse motion sources.
(b) \method follows a two-stage training framework: Stage I pretrains a language-conditioned Motion Expert, which is integrated with Video and Action Experts during Stage II robot post-training.
(c) Heterogeneous motion sources are unified into a shared root-free 63D motion space, enabling the learned motion prior to transfer to real-world TianGong 3.0 whole-body tasks.
\textbf{See our project page:} \textcolor{blue}{\url{https://zbzyjya.github.io/WholeBodyWAM/}}
}
\label{fig:teaser}
\end{strip}


\input{ICRA_Submission/1-abstract}

\input{ICRA_Submission/2-intro}

\input{ICRA_Submission/3-related}

\input{ICRA_Submission/4-method}

\input{ICRA_Submission/5-exp}
\input{ICRA_Submission/6-con}

\bibliographystyle{IEEEtran}
\bibliography{references_icra2027}


\appendix
\input{appendix}

\end{document}

%% file: ICRA_Submission/0-macro.tex
\usepackage{amsmath,amssymb,amsfonts}
\usepackage{graphicx}
\usepackage{booktabs}
\usepackage{multirow}
\usepackage{array}
\usepackage{tabularx}
\usepackage{microtype}
\usepackage{cite}
\usepackage[draft,bookmarks=false]{hyperref}
\usepackage{url}
\usepackage{xspace}
\usepackage{placeins}
\usepackage{makecell}
\usepackage{cuted}    
\usepackage{capt-of}  

\newcommand{\method}{\textsc{WholeBodyWAM}\xspace}

%% file: ICRA_Submission/1-abstract.tex
\begin{abstract}
Humanoid whole-body manipulation requires coordinated whole-body dynamics, yet large-scale trajectories from a target robot are expensive to collect and difficult to scale. In contrast, whole-body motion from human and humanoid sources is abundantly available, although such data cannot be directly used as embodiment-specific robot actions. This work asks whether these scalable motion resources can instead provide a transferable predictive prior for humanoid world-action modeling. We introduce \method, a humanoid world-action model that learns whole-body dynamics from large-scale heterogeneous motion before target-robot training. We curate UniMotion-4K, a motion corpus spanning more than 4K hours from human videos, native 3D motion datasets, and heterogeneous humanoid platforms, and canonicalize these diverse sources into a unified motion space. A language-conditioned Motion Expert is then pretrained to predict future whole-body motion without target-robot action supervision. During robot post-training, the pretrained Motion Expert is integrated with Video and Action Experts through asymmetric Mixture-of-Transformers (MoT) attention, enabling predictive scene dynamics and whole-body motion to jointly inform embodiment-specific action generation. Experiments show that \method consistently benefits from increased motion-pretraining scale, improves future-motion prediction and downstream task performance, and transfers effectively to real-world humanoid manipulation. Moreover, the pretrained motion prior substantially improves data efficiency under limited target-robot demonstrations.
\end{abstract}

%% file: ICRA_Submission/2-intro.tex
\section{Introduction}

Humanoid whole-body control requires coordinated posture transitions, locomotion when necessary, and object interaction over extended task horizons. A growing line of work applies foundation models to humanoids through hierarchical control architectures, where a high-level model generates task-conditioned commands that are executed by a reinforcement-learning-based whole-body controller~\cite{chen2025hand,humanoidvla2025}. Within this framework, Vision-Language-Action models (VLAs) have been increasingly explored for mapping visual observations and language instructions to high-level control commands~\cite{grootn1_2025,psi0_2026,wholebodyvla2025,hex2026}. In contrast, World Action Models (WAMs), which explicitly model future dynamics to provide predictive foresight for action generation~\cite{unipi2023,vpp2025,fastwam2026,tau0wm2026}, remain relatively underexplored for humanoid whole-body control~\cite{motionwam2026}. This motivates investigating predictive world-action modeling for the temporally extended and highly coordinated behaviors required by humanoids.

Training such foundation models, however, typically requires large-scale data~\cite{robomind2025,robomind2_2025}, while collecting humanoid demonstrations is expensive and labor-intensive. Learning predictive whole-body dynamics solely from target-robot data is therefore difficult to scale. In contrast, large-scale whole-body motion is readily available, providing diverse examples of posture transitions, coordinated movement, locomotion, and temporally extended behavior. Although these data do not contain executable commands for the target humanoid, they provide scalable supervision for learning task-conditioned motion priors before target-robot training. We therefore curate UniMotion-4K, comprising more than 4K hours of whole-body motion from 11 heterogeneous sources, including human videos, native 3D motion datasets, and heterogeneous humanoid platforms. To make these sources jointly usable, we develop a unified processing pipeline that canonicalizes heterogeneous body models, joint definitions, and state parameterizations into a shared motion representation.

To exploit these scalable motion resources, we propose \method, a Video-Motion-Action world-action model that first learns a predictive whole-body prior from large-scale motion data and subsequently grounds it into target-robot control. Our key insight is that existing WAMs primarily emphasize future visual dynamics whereas humanoid control also critically depends on the temporal evolution of the robot's own articulated body. For a high-degree-of-freedom humanoid, the torso, arms, and lower body must remain coordinated throughout task execution. The articulated body is therefore itself a control-relevant component of the evolving world state, making future whole-body motion a natural predictive signal for humanoid WAMs.

As illustrated in Fig.~\ref{fig:teaser}, \method adopts a two-stage training framework. In Stage I, a language-conditioned Motion Expert is pretrained on UniMotion-4K to predict task-conditioned future motion without target-robot action supervision. In Stage II, the pretrained Motion Expert is integrated with Video and Action Experts through layer-wise Mixture-of-Transformers attention~\cite{mot2025}, allowing predictive scene dynamics and whole-body motion to jointly inform embodiment-specific action generation. Future visual dynamics provide predictive supervision during training, while future visual tokens and video decoding are omitted at deployment. Consequently, \method directly predicts future whole-body motion and executable action chunks in closed loop, retaining predictive foresight while avoiding the additional inference overhead of explicit future-video generation.
Our contributions are threefold:
\begin{itemize}
    \item We curate UniMotion-4K, a 4K-hour whole-body motion corpus spanning human videos, native 3D motion datasets, and heterogeneous humanoid platforms, together with a unified pipeline that maps these diverse sources into a shared motion space.

    \item We propose \method, a Video-Motion-Action world-action model that pretrains a task-conditioned Motion Expert on UniMotion-4K and integrates it with Video and Action Experts to transfer predictive whole-body priors to target-robot control.

    \item Extensive real-world experiments demonstrate strong performance on challenging humanoid whole-body manipulation tasks and show that scaling motion pretraining consistently improves future-motion prediction, downstream control, and target-robot data efficiency.
\end{itemize}

%% file: ICRA_Submission/3-related.tex
\section{Related Work}

\subsection{Humanoid Whole-Body Policies}

Recent humanoid whole-body policies increasingly adopt hierarchical architectures, where foundation models generate high-level task-conditioned commands and reinforcement-learning-based whole-body controllers execute them~\cite{bai2025towards}. Most foundation-model-based humanoid policies follow the Vision-Language-Action (VLA) paradigm. Humanoid-VLA~\cite{humanoidvla2025} integrates language, egocentric perception, and motion generation for general humanoid control, while the GR00T N1 series~\cite{grootn1_2025}, $\Psi_0$~\cite{psi0_2026}, and WholeBodyVLA~\cite{wholebodyvla2025} explore scalable pretraining and generalist whole-body loco-manipulation. HEX~\cite{hex2026} further exploits heterogeneous robot trajectories through humanoid-aligned proprioceptive modeling, while HAF~\cite{gu2026haf} adapts generalist VLAs to humanoids through hierarchical action generation and reinforcement-learning-based policy refinement. Collectively, these works demonstrate the potential of foundation models for humanoid whole-body control, but primarily rely on current observations without explicitly modeling future dynamics.

World Action Models (WAMs) couple action generation with explicit prediction of future dynamics, providing policies with foresight beyond reactive control. Existing approaches either condition action prediction on generated future representations~\cite{unipi2023,vpp2025} or jointly model future dynamics and actions within a unified generative framework~\cite{fastwam2026,dreamzero2026,cosmospolicy2026}. Despite this progress, WAMs remain relatively underexplored for humanoid whole-body control. MotionWAM~\cite{motionwam2026} introduces world-action modeling for real-time humanoid loco-manipulation by conditioning motion prediction on video-world-model features. In contrast, \method focuses on acquiring predictive body knowledge at scale: it pretrains future whole-body dynamics from heterogeneous motion data and subsequently grounds the learned motion prior into target-robot action generation. In this way, \method treats future articulated-body evolution as an explicit predictive modality alongside scene dynamics for humanoid world-action modeling.

\subsection{Large-Scale Motion Modeling}

Large-scale human motion datasets provide rich supervision for learning articulated body dynamics. HumanML3D~\cite{humanml3d2022} pairs 3D motion with language, Motion-X~\cite{motionx2023} scales to diverse multimodal whole-body motion, and BEDLAM~\cite{bedlam2023} provides accurately annotated SMPL-X~\cite{smplx2019} motion from rendered videos. For humanoid learning, SONIC~\cite{sonic2025} demonstrates the benefits of scaling motion data and model capacity for whole-body tracking, while BONES-SEED~\cite{bonesseed2026} provides large-scale annotated motion with humanoid-compatible retargeting.
Recent generative motion models further demonstrate that complex temporal dynamics can be learned from diverse motion corpora. CondMDI~\cite{condmdi2024} studies diffusion-based motion in-betweening, MMM~\cite{mmm2023} employs masked discrete motion modeling, and MotionBricks~\cite{motionbricks2026} scales latent motion generation to large motion collections and real-time control. In contrast to these works, which primarily treat motion generation or tracking as the end task, \method uses large-scale motion as transferable predictive supervision for world-action modeling, enabling learned whole-body dynamics to support downstream humanoid control.

%% file: ICRA_Submission/4-method.tex
\section{Method}

\input{ICRA_Submission/4-1-overview}

\input{ICRA_Submission/4-2-motion_space}

\input{ICRA_Submission/4-3-motion_corpus}

\input{ICRA_Submission/4-4-pretrain}

\input{ICRA_Submission/4-5-posttrain}

\input{ICRA_Submission/4-6-closed-loop}

%% file: ICRA_Submission/4-1-overview.tex
\method learns a scalable whole-body motion prior and transfers it to target-robot control. We canonicalize heterogeneous human and humanoid motion into a unified root-free 63D space and construct UniMotion-4K. As shown in Fig.~\ref{fig:model}, Stage I pretrains a Motion Expert for task-conditioned future motion prediction, while Stage II integrates it with Video and Action Experts through a unified Video--Motion--Action world-action architecture with asymmetric cross-modal attention.

%% file: ICRA_Submission/4-2-motion_space.tex
\subsection{UniMotion-4K: Data Composition and Processing}

\subsubsection{Data Composition and Motion Unification}

UniMotion-4K integrates heterogeneous human and humanoid motion with different coordinate systems, body models, joint definitions, and state parameterizations. We canonicalize these sources into a unified root-free 63D motion space, where each frame represents the local rotations of 21 joints using 3D axis-angle vectors. Global root translation, root orientation, and body shape are excluded, retaining only articulated body configurations. This shared representation enables cross-embodiment motion modeling, while embodiment-specific global movement and executable control remain in the native action space.

\paragraph{Exocentric Human Videos}
We use HowTo100M~\cite{howto100m2019} as the primary source of large-scale unconstrained human videos. After person detection, tracking, and quality filtering, we recover temporally coherent 3D human motion using GVHMR~\cite{gvhmr2024}. The estimated SMPL-X~\cite{smplx2019} parameters are converted into the unified representation by retaining the local rotations of the selected 21 joints.

\paragraph{Egocentric Human Data}
For datasets with synchronized 3D body-motion annotations, such as EgoDex~\cite{egodex2026}, we directly convert the provided body parameters without using RGB videos. We additionally include Ego-Exo4D~\cite{egoexo4d2024}; when explicit 3D body motion is unavailable, we first recover full-body motion from the available visual and motion cues and then convert it into the same 63D representation.

\paragraph{Native 3D Human Motion}
We incorporate native 3D human motion datasets including HumanML3D~\cite{humanml3d2022}, Motion-X~\cite{motionx2023}, AMASS~\cite{amass2019}, BONES-SEED~\cite{bonesseed2026} (Motion Data by Bones Studio, \url{https://bones.studio/}), and BEDLAM~\cite{bedlam2023}. Their heterogeneous body models and joint conventions are canonicalized to a common 21-joint definition, after which the corresponding local axis-angle rotations are retained to form 63D motion sequences.

\paragraph{Heterogeneous Humanoid Motion}
Native humanoid datasets provide embodiment-specific joint states rather than human-parametric body representations. To map them into the unified motion space, we construct an embodiment-specific adapter for each humanoid. Specifically, we use GMR~\cite{gmr2025} to retarget parametric human motion to embodiment $e$, producing paired native robot joint states and corresponding unified 63D motions. These pairs supervise an adapter
\begin{equation}
    G_e:\mathbb{R}^{d_e}\rightarrow\mathbb{R}^{63},
\end{equation}
where $d_e$ denotes the native joint-state dimension of embodiment $e$. We implement $G_e$ as a three-layer MLP with two 128-dimensional hidden layers and ReLU activations, with input and output z-score normalization. Once trained, $G_e$ converts native humanoid datasets such as HIW-500~\cite{hiw500_2026} and UnifoLM-WBT~\cite{unifolmwbt2026} into unified 63D motion sequences. For the target robot, its own adapter maps proprioceptive states into the unified motion space during Stage II and deployment, and remains frozen throughout Stage II.

\begin{figure*}[t]
\centering
\includegraphics[width=0.94\textwidth]{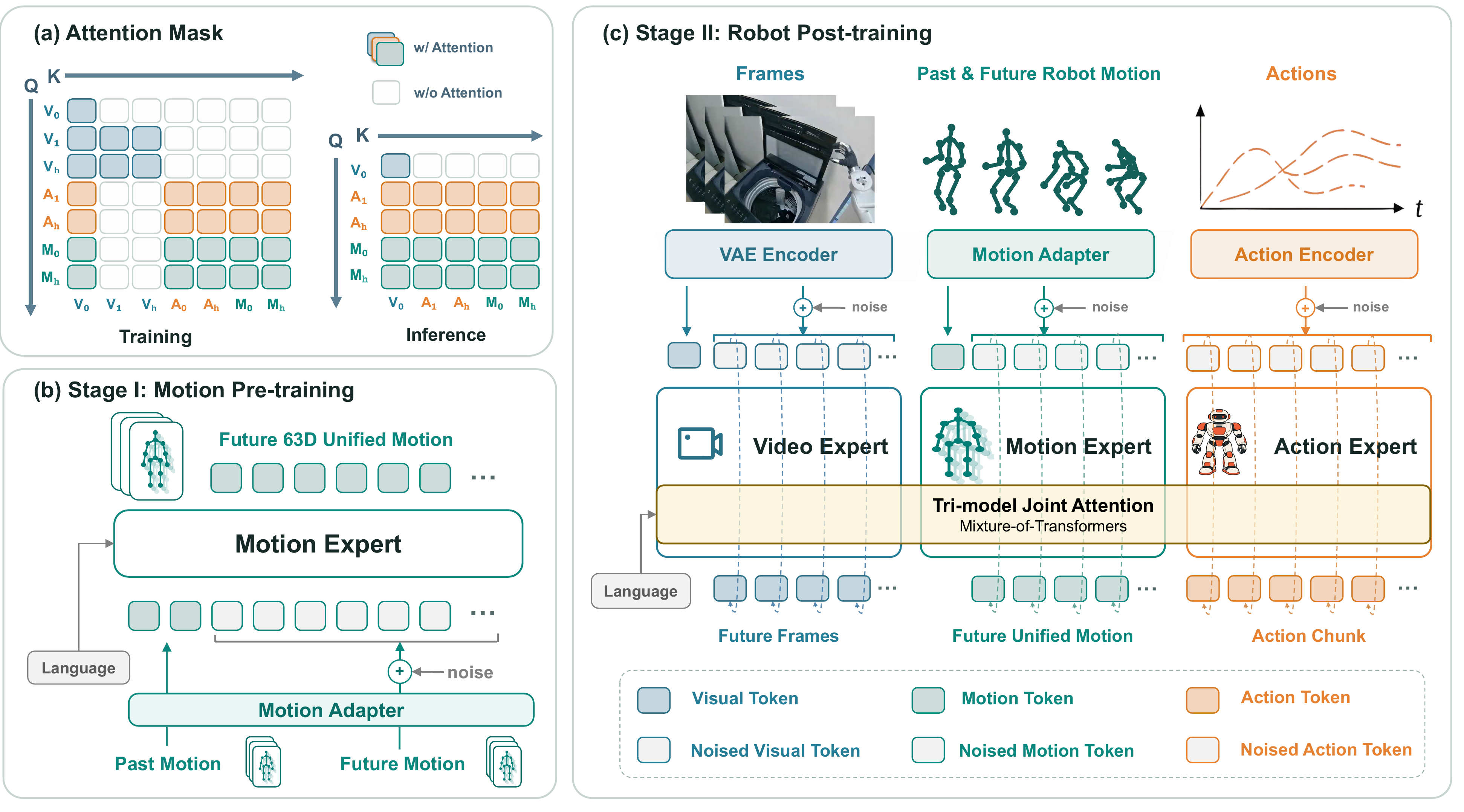}
\vskip -0.12in
\caption{\textbf{Model architecture and two-stage training paradigm of \method.}
(a) Asymmetric attention masks regulate cross-modal information flow during Stage II training and inference.
(b) Stage I pretrains a language-conditioned Motion Expert for future 63D whole-body motion prediction.
(c) Stage II jointly post-trains Video, Motion, and Action Experts through layer-wise Mixture-of-Transformers attention. 
A subset of the predicted action chunk directly controls the manipulators, head, and waist, while the remaining high-level commands are passed to an RL-based whole-body controller to generate lower-body actions for locomotion and balance.
}
\label{fig:model}
\end{figure*}

%% file: ICRA_Submission/4-3-motion_corpus.tex
\subsubsection{Data Processing}
\label{sec:motion_dataset}

Structured motion datasets provide accurate whole-body supervision but remain limited in scale and behavioral diversity. We therefore construct UniMotion-4K from heterogeneous sources, including Internet videos, egocentric human data, native 3D human motion, and humanoid motion, all converted into the unified 63D representation.

\paragraph{Dataset Construction}
We begin with more than 30K hours of heterogeneous source data. Internet videos are segmented into temporally coherent human-centric clips and filtered for severe body truncation or occlusion, poor temporal continuity, and insufficient body motion. Retained clips undergo 3D motion recovery, while sequences with failed reconstruction, non-finite parameters, implausibly large joint rotations, abrupt frame-to-frame pose discontinuities, or unstable root motion are discarded. Other data sources are processed using the corresponding conversion procedures described above. After filtering and canonicalization, UniMotion-4K contains approximately 1.19M variable-length sequences, totaling 444M motion frames and more than 4.1K hours of whole-body motion.

\paragraph{Motion-Centric Language Annotation}
For video-derived sequences, we generate language annotations using Qwen2.5-VL-3B-Instruct~\cite{qwen25vl2025} with motion-focused prompts emphasizing body posture, limb coordination, locomotion, temporal transitions, and object interactions rather than static scene appearance. Existing semantic annotations from structured motion datasets are retained when available. The resulting motion--language pairs are used for language-conditioned Stage I pretraining.

%% file: ICRA_Submission/4-4-pretrain.tex
\subsection{Stage I: Scalable Motion Pretraining}
\label{sec:stage1}

As shown in Fig.~\ref{fig:model} (b), Stage I pretrains the Motion Expert on UniMotion-4K to learn task-conditioned future whole-body dynamics from language--motion pairs, without target-robot action supervision.
The Motion Expert consists of 30 transformer blocks with 512-dimensional tokens and a $4\times$ feed-forward expansion. Each 63D motion frame is projected into the token space. We sample 48-frame windows from variable-length sequences, using 16 frames as motion history $\mathbf{M}^{\mathrm{hist}}$ and the subsequent 32 frames as the future prediction target $\mathbf{M}^{\mathrm{fut}}$. Historical and future motion share the same projection and are distinguished by temporal positional embeddings, token-type embeddings, and flow-timestep conditioning. The language instruction $\ell$ is represented by precomputed embeddings from the frozen Wan2.2 text encoder~\cite{wan2025}.

We formulate future-motion prediction using continuous flow matching~\cite{flowmatching2023}. Let $\boldsymbol{\epsilon}\sim\mathcal{N}(0,\mathbf{I})$ denote Gaussian noise and $\sigma\in[0,1]$ the flow timestep. The noisy future motion $\mathbf{X}_{\sigma}$ and target velocity $\mathbf{u}^{*}$ are defined as
\begin{equation}
    \mathbf{X}_{\sigma}
    =
    (1-\sigma)\mathbf{M}^{\mathrm{fut}}
    +
    \sigma\boldsymbol{\epsilon},
    \qquad
    \mathbf{u}^{*}
    =
    \boldsymbol{\epsilon}-\mathbf{M}^{\mathrm{fut}}.
\end{equation}
The Motion Expert $F_m$ predicts the velocity field from motion history, noisy future motion, language, and flow timestep:
\begin{equation}
    \mathcal{L}_{m}
    =
    \mathbb{E}
    \left[
    \left\|
    F_m(\mathbf{M}^{\mathrm{hist}},\mathbf{X}_{\sigma},\ell,\sigma)
    -
    \mathbf{u}^{*}
    \right\|_2^2
    \right].
\end{equation}

At inference, future motion is initialized from Gaussian noise and generated by integrating the learned velocity field from $\sigma=1$ to $\sigma=0$. Since Stage I requires neither target-robot actions nor target-robot demonstrations, the Motion Expert learns a transferable predictive prior from UniMotion-4K before Stage II grounding.

%% file: ICRA_Submission/4-5-posttrain.tex
\subsection{Stage II: Video-Motion-Action Post-training}
\label{sec:stage2}

Stage II grounds the pretrained motion prior into target-robot control by jointly post-training Video, Motion, and Action Experts on target-robot demonstrations.

\subsubsection{Video-Motion-Action Architecture}
\label{sec:vma_arch}

As shown in Fig.~\ref{fig:model} (c), given the current visual observation $\mathbf{I}_t$, recent motion history $\mathbf{M}^{\mathrm{hist}}_t$, and language instruction $\ell$, \method jointly predicts future scene dynamics, future whole-body motion, and executable robot actions. The Video Expert is initialized from Wan2.2-TI2V-5B~\cite{wan2025}, while the Motion Expert is initialized from Stage I. The Action Expert uses a 30-layer DiT for action generation.
The three Experts interact layer by layer through Mixture-of-Transformers (MoT) attention~\cite{mot2025}. Their modality-specific queries, keys, and values participate in joint attention while retaining separate expert streams, enabling predictive information to be exchanged across visual, motion, and action representations.

\subsubsection{Asymmetric Video-Motion-Action Attention}
\label{sec:asym_mask}

As shown in Fig.~\ref{fig:model}(a), Stage II adopts asymmetric attention to prevent future visual information from leaking into motion and action prediction. Let $V_0$ denote the visual tokens of the current observation and $V_{1:H}$ the future visual tokens used for prediction. During training, the Motion and Action streams can attend to $V_0$ and to each other, but cannot access $V_{1:H}$. Future visual latents therefore provide predictive supervision without privileged future information. At inference, $V_{1:H}$ is omitted entirely, while $V_0$ continues to condition motion and action generation.

\subsubsection{Joint Post-training Objective}
\label{sec:joint_posttraining}

The Video, Motion, and Action Experts are jointly optimized to predict future visual latents, future whole-body motion, and target-robot action chunks, respectively. The Motion Expert follows the flow-matching formulation introduced in Stage I. Let $\mathbf{Z}^{\mathrm{fut}}$ denote the future visual latents and $\mathbf{A}$ the target action chunk. For the Video and Action Experts, we define
\begin{equation}
    \mathbf{Y}_v=\mathbf{Z}^{\mathrm{fut}},
    \qquad
    \mathbf{Y}_a=\mathbf{A},
\end{equation}
where $\mathbf{Y}_k$ denotes the prediction target of modality $k$. Their flow-matching objectives are
\begin{equation}
\begin{aligned}
    \mathcal{L}_k
    &=
    \mathbb{E}
    \left[
    \left\|
    F_k(\mathbf{X}_{\sigma}^{k};\mathcal{C}_k,\sigma)
    -
    (\boldsymbol{\epsilon}_k-\mathbf{Y}_k)
    \right\|_2^2
    \right], \\
    \mathbf{X}_{\sigma}^{k}
    &=
    (1-\sigma)\mathbf{Y}_k
    +
    \sigma\boldsymbol{\epsilon}_k,
    \qquad
    k\in\{v,a\},
\end{aligned}
\end{equation}
where $F_k$ denotes the corresponding Video or Action Expert, $\boldsymbol{\epsilon}_k$ is Gaussian noise, and $\mathcal{C}_k$ denotes the conditioning available to each Expert under the asymmetric attention mask. The joint post-training objective is
\begin{equation}
    \mathcal{L}_{\mathrm{robot}}
    =
    \lambda_v\mathcal{L}_v
    +
    \lambda_m\mathcal{L}_m
    +
    \lambda_a\mathcal{L}_a,
\end{equation}
where $\lambda_v=\lambda_m=\lambda_a=1$. The pretrained Motion Expert remains trainable throughout Stage II. Stage I and Stage II use the same 63D motion representation and prediction horizon, enabling direct transfer of the pretrained Motion Expert.

%% file: ICRA_Submission/4-6-closed-loop.tex
\subsection{Closed-Loop Humanoid Control}
\label{sec:control}

At deployment, real-robot policy inference is performed on a single NVIDIA RTX 5090 GPU. The robot receives the current onboard visual observation $\mathbf{I}_t$ and proprioceptive state $\mathbf{q}_t$. The embodiment adapter $G_e$ maps $\mathbf{q}_t$ into the unified motion space, and the latest 16 motion states form the motion history $\mathbf{M}^{\mathrm{hist}}_t$. Given $\mathbf{I}_t$, $\mathbf{M}^{\mathrm{hist}}_t$, and language instruction $\ell$, the joint model $\mathcal{F}_{\theta}$ predicts future whole-body motion and an executable action chunk:
\begin{equation}
    (\hat{\mathbf{M}}^{\mathrm{fut}}_t,\hat{\mathbf{A}}_t)
    =
    \mathcal{F}_{\theta}
    (\mathbf{I}_t,\mathbf{M}^{\mathrm{hist}}_t,\ell).
\end{equation}
where $\hat{\mathbf{M}}^{\mathrm{fut}}_t$ is the predicted future motion used only as a predictive representation, while $\hat{\mathbf{A}}_t$ contains embodiment-specific commands in the native action space.

We adopt chunked closed-loop execution. At each policy query, future motion and action are sampled using four Euler steps. Arm, hand, head, and waist commands are executed directly, while leg commands are generated by the RL-based whole-body controller for locomotion and balance. The 32-step action chunk is executed at 30\,Hz, after which new visual and proprioceptive observations are acquired and the policy replans. Future visual latents are used only for Stage II supervision and omitted at deployment, avoiding future-video generation and decoding.

%% file: ICRA_Submission/5-exp.tex
\section{Experiments}

We evaluate \method on real-world humanoid whole-body tasks to assess its control performance, generalization, and the effects of motion pretraining. Specifically, we investigate three questions:
\textbf{RQ1:} How does \method compare with representative VLA and WAM baselines on real-world whole-body control?
\textbf{RQ2:} How well does \method generalize to variations in task conditions, such as unseen objects and spatial configurations?
\textbf{RQ3:} How do motion modeling, pretraining scale, and target-robot data availability affect downstream performance?

\input{ICRA_Submission/5-1-setup}

\begin{figure*}[t]
    \centering
    \includegraphics[width=0.96\textwidth]{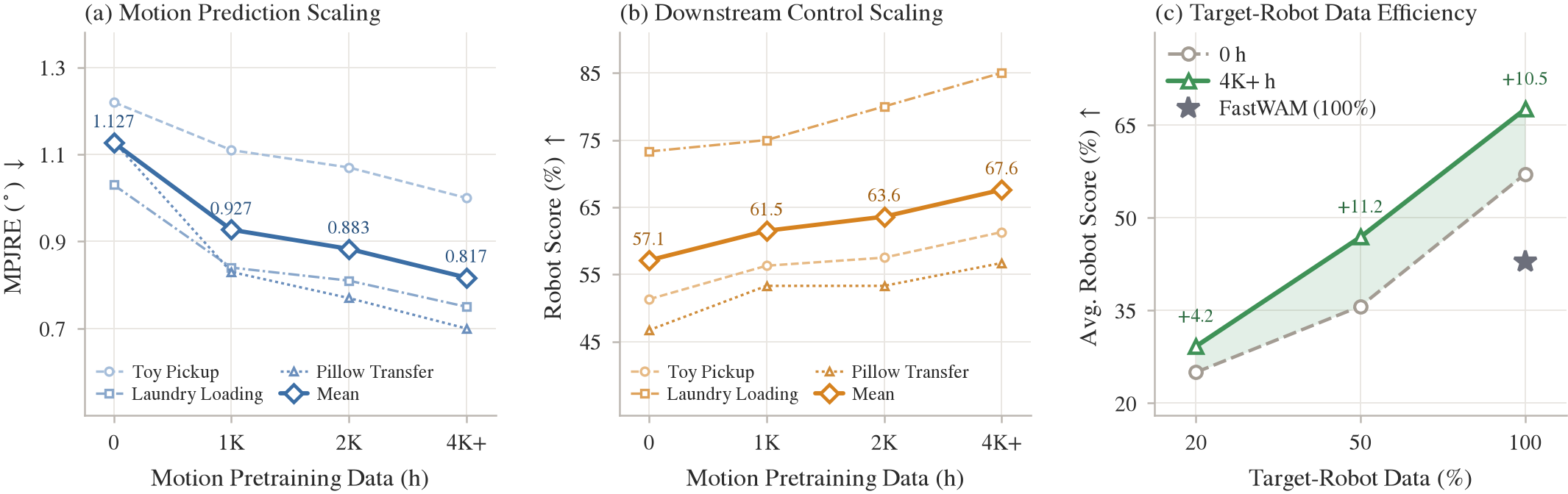}
    \vskip -0.1in
    \caption{\textbf{Scaling analysis of motion pretraining.}
    (a) Increasing motion pretraining data consistently reduces future whole-body motion prediction error across three representative tasks.
    (b) Larger-scale motion pretraining improves downstream real-robot task scores.
    (c) Motion pretraining improves target-robot data efficiency; with 50\% of the target-robot demonstrations, the 4K+ h model surpasses FastWAM trained on the full demonstration set.}
    \label{fig:motion_analysis}
\end{figure*}

\input{ICRA_Submission/5-2-real}
\input{ICRA_Submission/5-3-ablation}

%% file: ICRA_Submission/5-1-setup.tex
\subsection{Experimental Setup}

\begin{figure*}[t]
    \centering
    \includegraphics[width=0.98\textwidth]{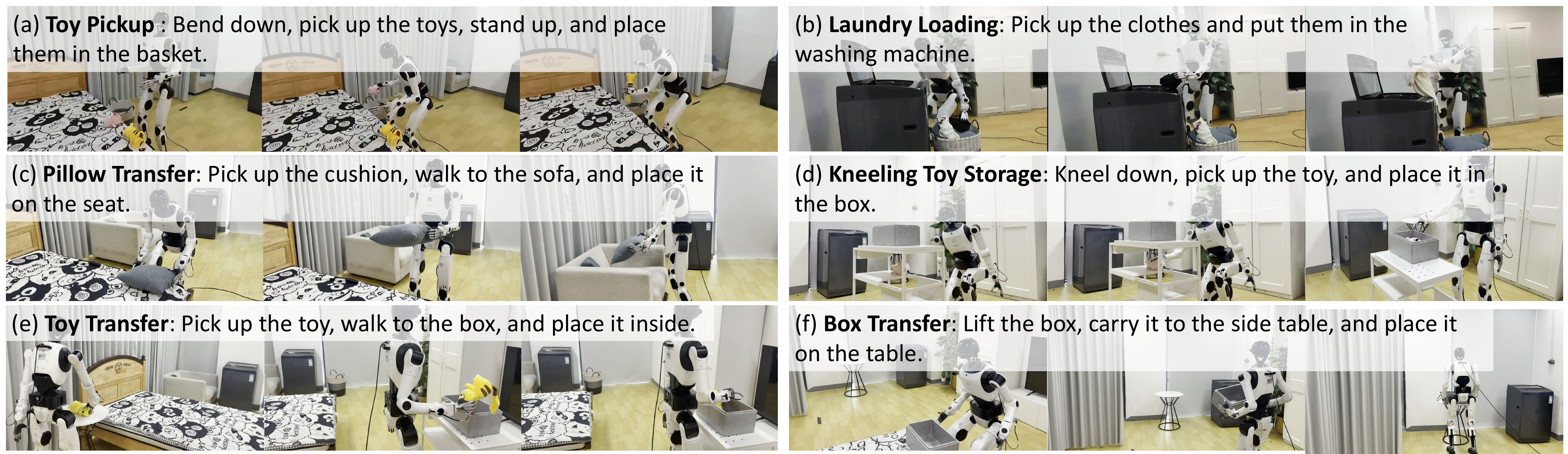}
    \vskip -0.1in
    \caption{\textbf{Real-world humanoid whole-body tasks.}
    The six tasks cover coordinated whole-body manipulation, posture transitions, and loco-manipulation.}
    \label{fig:tasks}
\end{figure*}

\begin{table*}
\centering
\caption{\textbf{Real-world performance on humanoid whole-body tasks.}
We report normalized task score (\%); the best results are bold.}
\label{tab:main_results}
\vskip -0.1in
\resizebox{0.98\textwidth}{!}{
\begin{tabular}{lccccccc}
\toprule
\textbf{Method}
&
\makecell[c]{\textbf{Toy}\\\textbf{Pickup}}
&
\makecell[c]{\textbf{Laundry}\\\textbf{Loading}}
&
\makecell[c]{\textbf{Pillow}\\\textbf{Transfer}}
&
\makecell[c]{\textbf{Kneeling Toy}\\\textbf{Storage}}
&
\makecell[c]{\textbf{Toy}\\\textbf{Transfer}}
&
\makecell[c]{\textbf{Box}\\\textbf{Transfer}}
&
\makecell[c]{\textbf{Avg.}}
\\
\midrule
FastWAM~\cite{fastwam2026}
& 43.8 & 51.7 & 33.3
& 66.7 & 23.3 & 30.0
& 41.5 \\
$\pi_{0.5}$~\cite{pi05_2025}
& 50.0 & \textbf{85.0} & 46.7
& 70.0 & 53.3 & 43.3
& 58.1 \\
GR00T N1.7~\cite{grootn1_2025}
& 56.3 & 60.0 & 53.3
& 81.7 & 60.0 & 53.3
& 60.8 \\
$\tau_0$-WM~\cite{tau0wm2026}
& 41.3 & 40.0 & 26.7
& 78.3 & 26.7 & 23.3
& 39.4 \\
\midrule
$\tau_0$-WM + Motion Pretraining
& \textbf{61.3} & 66.7 & 46.7
& 85.0 & 53.3 & 63.3
& 62.7 \\
\midrule
\method w/o Motion Pretraining
& 51.3 & 73.3 & 46.7
& 80.0 & 50.0 & 53.3
& 59.1 \\
\method w/o Motion$\rightarrow$Action
& 42.5 & 58.3 & 33.3
& 63.3 & 36.7 & 43.3
& 46.3 \\
\textbf{\method (Ours)}
& \textbf{61.3} & \textbf{85.0} & \textbf{56.7}
& \textbf{90.0} & \textbf{66.7} & \textbf{73.3}
& \textbf{72.2} \\
\bottomrule
\end{tabular}
}
\end{table*}

\paragraph{Robot Platform and Data Collection}
We conduct all real-robot experiments on the TianGong 3.0 full-sized humanoid equipped with an onboard egocentric RGB camera. Following~\cite{xu2025hacts,hex2026}, whole-body demonstrations are collected using an IMU for head control, kinematically matched master arms for arm and hand teleoperation, and a handheld joystick for locomotion and waist control. We collect 100 target-robot demonstrations per task. The policy operates in a 34-dimensional whole-body action space covering locomotion, head and waist motion, and upper-body manipulation. All methods share the same robot platform, action interface, and RL-based low-level controller, isolating performance differences to the high-level policy.

\paragraph{Evaluated Tasks and Evaluation Metric}
As shown in Fig.~\ref{fig:tasks}, we evaluate six real-world whole-body tasks: \textit{Toy Pickup}, \textit{Laundry Loading}, \textit{Pillow Transfer}, \textit{Kneeling Toy Storage}, \textit{Toy Transfer}, and \textit{Box Transfer}. All tasks require coordinated posture adjustment and upper-body manipulation, while the three transfer tasks additionally involve loco-manipulation.
We conduct 20 rollouts for each task. Rather than binary success rate, we use a robot task score based on completed subtasks, normalized by the maximum achievable score for each task. For future-motion prediction, we report mean per-joint rotation error (MPJRE, degrees), defined as the geodesic rotation error between predicted and ground-truth joint rotations, averaged over all 21 joints and 32 future frames.

\paragraph{Baselines}
We compare \method with FastWAM~\cite{fastwam2026}, $\pi_{0.5}$~\cite{pi05_2025}, GR00T N1.7~\cite{grootn1_2025}, and $\tau_0$-WM~\cite{tau0wm2026}. We also evaluate \method w/o Motion Pretraining, which retains the same Video-Motion-Action architecture but removes Stage I initialization. To examine transferability, we augment $\tau_0$-WM with the pretrained Motion Expert through layer-wise video-motion MoT interactions. Motion Pretraining denotes initialization from the Motion Expert pretrained on the full UniMotion-4K dataset. All methods use the same target-robot demonstrations and RL-based whole-body controller. For each baseline, we use the released implementation and its recommended training settings.

%% file: ICRA_Submission/5-2-real.tex
\subsection{Real-World Evaluation}
\label{sec:real_robot_results}

\noindent\textbf{WholeBodyWAM achieves strong real-world performance.}
To address \textbf{RQ1}, we compare \method with representative VLA and WAM baselines on six real-world whole-body tasks. As shown in Table~\ref{tab:main_results}, \method achieves the highest average task score of 72.2\%, outperforming the strongest baseline, GR00T N1.7, by 11.4 percentage points. It achieves or matches the best performance across all six tasks, demonstrating consistent gains across whole-body manipulation and loco-manipulation.

\begin{figure}[!t]
    \centering
    \includegraphics[width=\columnwidth]{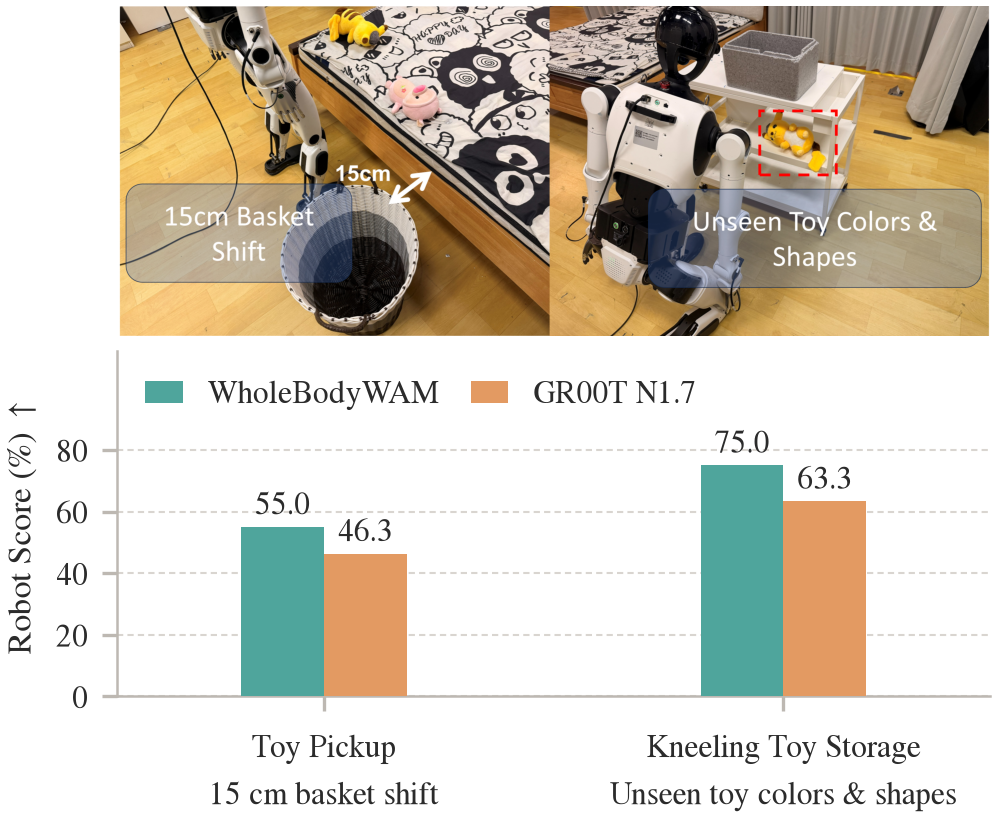}
    \vskip -0.1in
    \caption{\textbf{Generalization under controlled distribution shifts.}
    We evaluate spatial generalization with a 15\,cm basket shift in \textit{Toy Pickup} and object generalization with unseen toy colors and shapes in \textit{Kneeling Toy Storage}. Bars report normalized task scores (\%).}
    \label{fig:generalization}
\end{figure}

\subsection{Generalization Evaluation}
\label{sec:generalization}

\noindent\textbf{\method generalizes better to unseen task variations.}
To address \textbf{RQ2}, we evaluate two controlled distribution shifts beyond training: a 15\,cm basket shift in \textit{Toy Pickup} and unseen toy colors and shapes in \textit{Kneeling Toy Storage}. As shown in Fig.~\ref{fig:generalization}, \method outperforms GR00T N1.7 under both shifts, achieving 55.0\% vs.\ 46.3\% and 75.0\% vs.\ 63.3\%, respectively. It also exhibits smaller degradation from the in-distribution setting, indicating stronger robustness to spatial and object-level variations.

%% file: ICRA_Submission/5-3-ablation.tex
\subsection{Ablation and Analysis}
\label{sec:ablations}

We address \textbf{RQ3} through controlled studies of motion modeling, motion-pretraining scale, target-robot data efficiency, and inference cost.

\subsubsection{Effect of Motion Modeling}
\label{sec:motion_ablation}

As shown in Table~\ref{tab:main_results}, we ablate Stage I motion pretraining and Motion-to-Action information flow while keeping the remaining architecture unchanged. \method w/o Motion Pretraining removes Stage I initialization, whereas \method w/o Motion$\rightarrow$Action blocks Action queries from attending to Motion keys and values while retaining the pretrained Motion Expert and its prediction objective.

\noindent\textbf{Both motion pretraining and Motion-to-Action interaction are important.}
Removing Stage I pretraining reduces the average task score from 72.2\% to 59.1\%, while blocking Motion-to-Action attention further decreases it to 46.3\% despite retaining motion prediction supervision. These results demonstrate that both the pretrained motion prior and its direct interaction with action generation contribute substantially to \method.
We further integrate the pretrained Motion Expert into $\tau_0$-WM through layer-wise Video-Motion MoT interactions while preserving its original Video-Action interaction. This increases the average task score from 39.4\% to 62.7\%, providing evidence that the learned motion prior can transfer beyond the native \method architecture.

\subsubsection{Scaling with Motion Pretraining}
\label{sec:motion_scaling}

For the scaling study, we construct nested Stage I subsets containing approximately 1K, 2K, and 4K+ hours of UniMotion-4K, together with a 0\,h variant without motion pretraining. All pretrained variants use the same architecture and are trained for 100K Stage I steps under identical settings, followed by 20K Stage II steps on the same target-robot demonstrations. We evaluate \textit{Toy Pickup}, \textit{Laundry Loading}, and \textit{Pillow Transfer}, holding out 10 TianGong 3.0 trajectories per task for motion evaluation.

\noindent\textbf{Motion pretraining exhibits consistent scaling behavior.}
As shown in Fig.~\ref{fig:motion_analysis}(a-b), scaling Stage I data from 0 to 4K+ hours reduces average MPJRE from 1.127$^\circ$ to 0.817$^\circ$ (27.5\%) and improves the average real-robot task score from 57.1\% to 67.6\%. The aligned gains in motion prediction and control suggest that larger-scale motion pretraining yields more transferable whole-body dynamics.

\subsubsection{Target-Robot Data Efficiency}
\label{sec:data_efficiency}

We post-train \method using 20\%, 50\%, and 100\% of the target-robot demonstrations on three representative tasks. All variants use 20K Stage II steps, with identical demonstrations for models with and without motion pretraining under each data budget.

\noindent\textbf{Motion pretraining improves target-robot data efficiency.}
As shown in Fig.~\ref{fig:motion_analysis}(c), motion pretraining improves the average task score by 4.2, 11.2, and 10.5 percentage points at 20\%, 50\%, and 100\% data, respectively. With only 50\% of the demonstrations, the 4K+ hour pretrained model achieves 46.9\%, surpassing FastWAM trained on the full dataset (42.9\%). These results show that the learned motion prior reduces reliance on target-robot supervision.

\begin{figure}[!t]
    \centering
    \includegraphics[width=\columnwidth]{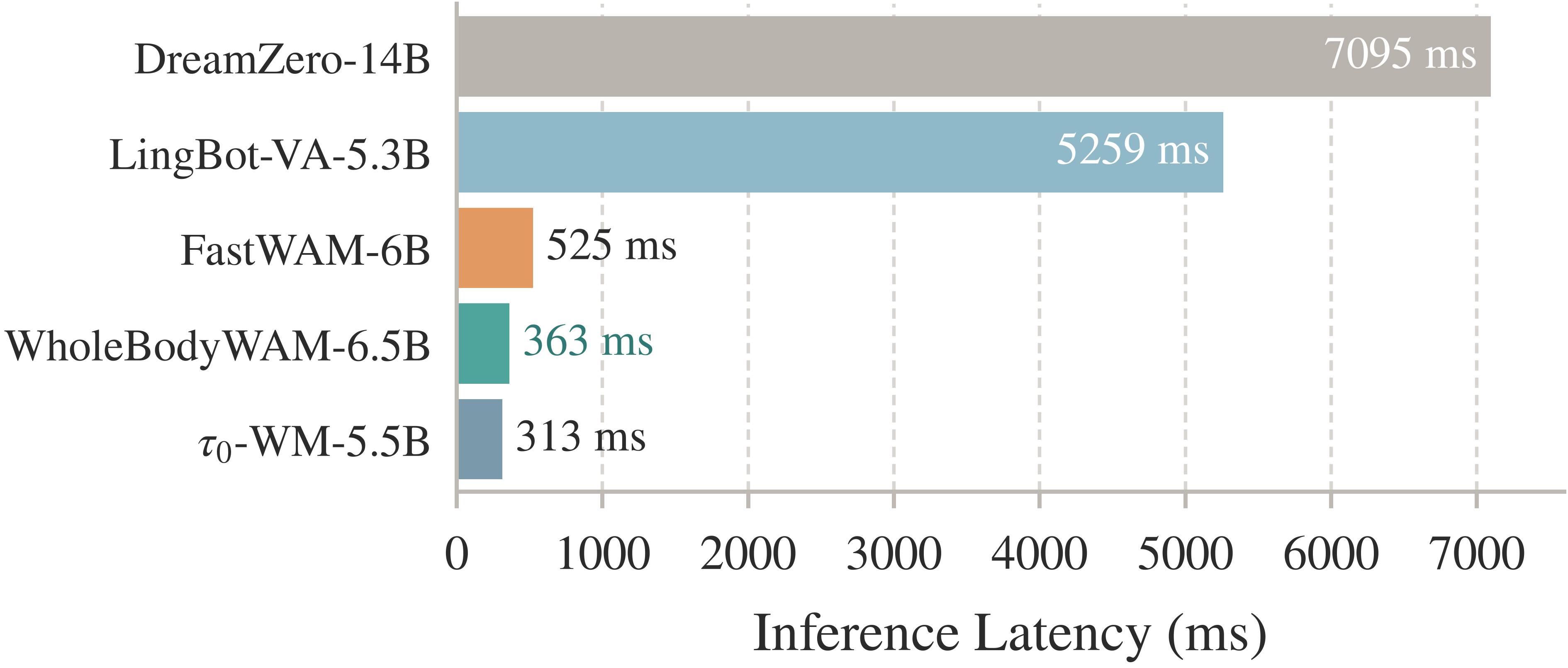}
    \vskip -0.1in
    \caption{End-to-end policy inference latency on an NVIDIA A100 GPU.}
    \label{fig:latency}
\end{figure}

\subsubsection{Inference Efficiency}
\label{sec:inference_efficiency}

For a controlled cross-method comparison, we benchmark end-to-end policy latency for all methods on a single NVIDIA A100 GPU using their respective deployment configurations.

\noindent\textbf{\method enables predictive modeling with efficient inference.}
As shown in Fig.~6, \method achieves 363\,ms end-to-end latency, $19.5\times$ and
$14.5\times$ faster than DreamZero~\cite{dreamzero2026} and
LingBot-VA~\cite{lingbotva2026}, respectively, and $1.45\times$
faster than FastWAM~\cite{fastwam2026}, while remaining close to
$\tau_0$-WM~\cite{tau0wm2026} (313\,ms).

%% file: ICRA_Submission/6-con.tex
\section{Conclusion}

We presented \method, a whole-body world-action model that leverages predictive whole-body motion as a scalable prior for control. By pretraining a Motion Expert on large-scale UniMotion-4K and integrating it with Video and Action Experts, \method incorporates explicit future body dynamics into action generation. Real-robot experiments across six whole-body tasks show that scaling motion pretraining consistently improves future-motion prediction, downstream control, and target-robot data efficiency. Ablations and cross-architecture transfer further validate the effectiveness and generality of the learned motion prior. These results demonstrate the value of scalable motion pretraining for humanoid world-action modeling.

%% file: appendix.tex

\setcounter{figure}{0}
\renewcommand{\thefigure}{A\arabic{figure}}

\setcounter{table}{0}
\renewcommand{\thetable}{A\arabic{table}}

\setcounter{equation}{0}
\renewcommand{\theequation}{A\arabic{equation}}

\subsection{UniMotion-4K Dataset and Processing}

\subsubsection{Dataset Composition}

UniMotion-4K is constructed from heterogeneous human and humanoid motion
sources. Some structured datasets partially overlap in their underlying
motions; for example, HumanML3D includes motions derived from AMASS.
We therefore report statistics over the retained sequences after
source-specific preprocessing, motion recovery, representation unification,
and quality filtering.

The final dataset contains 1,196,017 variable-length motion sequences,
approximately 444M motion frames, and over 4.1K hours of whole-body motion.
Here, a sequence denotes a variable-length motion segment retained after
preprocessing and is distinct from the fixed 48-frame windows sampled during
Stage I pretraining.

\subsubsection{Internet-Video Processing and Quality Control}

For Internet-video sources, we detect and track the primary human subject,
segment long videos into temporally coherent human-centric clips, and apply
quality control both before and after 3D motion recovery.

\paragraph{Video-Level Clip Filtering}
We remove clips with severe body truncation, unreliable pose observations,
fragmented tracks, or insufficient body movement. A keypoint is considered
visible when its confidence exceeds 0.30 and its location lies within the
image boundary. Candidate segments must have a valid-frame density of at
least 0.50 and contain at least 3.5\,s of temporally coherent motion.

We further distinguish full-body from partial-body observations using torso
and lower-body joint visibility. Clips containing only isolated hand or wrist
detections without reliable torso observations are rejected. To remove nearly
static segments, we require body-scale-normalized limb or torso displacement
to exceed 0.02. Predominantly static or talking segments are additionally
filtered during motion-centric language annotation.

\paragraph{3D Motion Recovery}
Clips passing the video-level filters are processed by 2D pose estimation and
temporally coherent 3D motion recovery using GVHMR. Samples for which
tracking or recovery fails to produce a valid motion sequence are discarded.

\paragraph{Motion-Level Quality Control}
Recovered sequences are further checked for numerical validity and temporal
consistency. We reject sequences shorter than 15 frames, sequences containing
NaN or Inf values, and motions with implausibly large joint rotations.
In particular, sequences containing absolute axis-angle values greater than
5.0 are removed.
\begin{table}[!t]
\centering
\caption{Composition of UniMotion-4K.}
\label{tab:appendix_dataset}
\vskip -0.1in
\small
\begin{tabular}{lr}
\toprule
\textbf{Dataset} & \textbf{\# Sequences} \\
\midrule
HowTo100M       & 501,746 \\
EgoDex          & 314,839 \\
HIW-500         & 176,768 \\
BONES-SEED      & 135,246 \\
AMASS           & 25,366 \\
Motion-X        & 9,962 \\
BEDLAM          & 9,949 \\
UnifoLM-WBT     & 8,989 \\
HAA500          & 6,944 \\
HumanML3D       & 5,826 \\
EgoExo4D        & 382 \\
\midrule
\textbf{Total}  & \textbf{1,196,017} \\
\bottomrule
\end{tabular}
\vspace{-0.4em}
\end{table}
We also measure frame-to-frame changes in articulated pose and root
translation to detect unstable reconstructions. Sequences with severe pose
discontinuities or excessive root-translation jumps are discarded, primarily
removing failures caused by tracking switches, unstable 3D reconstruction,
or physically implausible motion.

Only sequences passing both video-level filtering and post-recovery quality
control are retained in UniMotion-4K.

\FloatBarrier

\subsection{Architecture and Training Details}

This section provides additional architectural and training details of
\method, including modality-specific tokenization, flow-timestep sampling,
layer-wise cross-modal interaction, attention connectivity, and optimization
settings.

\subsubsection{Stage I Motion Tokenization and Conditioning}

Each 63D motion frame is represented as a single motion token.
A motion frame $\mathbf{m}_t\in\mathbb{R}^{63}$ is projected into the
Motion Expert token space through
\begin{equation}
\mathbf{h}_t^m
=
W_{m,2}
\,\mathrm{GELU}
\left(
W_{m,1}\mathbf{m}_t+b_{m,1}
\right)
+b_{m,2},
\end{equation}
where both the intermediate and output dimensions are 512.
Historical and future frames share the same projection network and are
distinguished by temporal positional embeddings and history/future type
embeddings.

Each Stage I sample contains 16 historical frames and 32 future frames.
The historical sequence remains clean and serves as conditioning, while only
the future sequence is perturbed by the flow process described in Sec.~III.B.
Language embeddings are precomputed using the frozen Wan2.2 text encoder and
condition the Motion Expert throughout Stage I training.

\subsubsection{Stage II Flow-Matching Details}

During Stage II, the Video, Motion, and Action streams use independently
sampled flow timesteps. For each modality $x\in\{v,m,a\}$, we sample
$u_x\sim\mathcal{U}(0,1)$ and apply timestep shifting
\begin{equation}
\sigma_x
=
\phi_{s_x}(u_x),
\qquad
\phi_s(u)
=
\frac{s\,u}
{1+(s-1)u}.
\end{equation}
We set $s_v=s_a=5$ for the Video and Action streams and $s_m=1$ for the
Motion stream.

Let $\mathbf{X}^v$, $\mathbf{X}^m$, and $\mathbf{X}^a$ denote the future
visual latents, future motion sequence, and action chunk, respectively.
For modality $x$, the noisy target and corresponding velocity target are
\begin{equation}
\mathbf{X}_{\sigma_x}^{x}
=
(1-\sigma_x)\mathbf{X}^{x}
+
\sigma_x\boldsymbol{\epsilon}_x,
\qquad
\mathbf{u}_x^{*}
=
\boldsymbol{\epsilon}_x-\mathbf{X}^{x},
\end{equation}
where $\boldsymbol{\epsilon}_x\sim\mathcal{N}(0,\mathbf{I})$.
The three streams are jointly optimized using the flow-matching objective
described in Sec.~III.C.

For the Motion stream, only the 32 future frames are noised, while the
16 historical frames remain clean conditioning inputs. For the Video stream,
the current-frame latent remains clean and serves as the visual condition.
The Action stream consists of 32 action tokens, each corresponding to one
34-dimensional robot action vector, projected as
\begin{equation}
\mathbf{h}_t^a
=
W_a\mathbf{a}_t+b_a,
\qquad
W_a:\mathbb{R}^{34}\rightarrow\mathbb{R}^{1024}.
\end{equation}
Unlike the Motion stream, the Action stream does not use history/future type
embeddings.

The Video stream follows the temporal subsampling of the video backbone.
With an action-to-video frequency ratio of 4, a 32-step action horizon
corresponds to 8 future video frames, together with the current conditioning
frame.

\subsubsection{Layer-Wise Mixture-of-Transformers Interaction}

The Video, Motion, and Action Experts retain modality-specific parameters
throughout the network while interacting through layer-wise
Mixture-of-Transformers attention. At layer $l$, for Expert
$x\in\{v,m,a\}$,
\begin{align}
Q_x^{(l)} &= H_x^{(l)}W_{Q,x}^{(l)},\\
K_x^{(l)} &= H_x^{(l)}W_{K,x}^{(l)},\\
V_x^{(l)} &= H_x^{(l)}W_{V,x}^{(l)}.
\end{align}
Each Expert maintains independent query, key, value, and output projections,
while its projected features participate in a common attention space.
Query and key features are normalized and equipped with rotary positional
embeddings before attention.

For query stream $x$, keys and values from the modalities permitted by its
attention mask are concatenated along the token dimension:
\begin{align}
K_{\mathcal{S}_x}^{(l)}
&=
\operatorname{Concat}
\left(
\{K_y^{(l)}:y\in\mathcal{S}_x\}
\right),\\
V_{\mathcal{S}_x}^{(l)}
&=
\operatorname{Concat}
\left(
\{V_y^{(l)}:y\in\mathcal{S}_x\}
\right),
\end{align}
where $\mathcal{S}_x$ denotes the set of modalities visible to query stream
$x$. Mixed attention is computed as
\begin{equation}
O_x^{(l)}
=
\operatorname{Softmax}
\left(
\frac{
Q_x^{(l)}
(K_{\mathcal{S}_x}^{(l)})^\top
}{\sqrt{d}}
+
\mathcal{M}_x
\right)
V_{\mathcal{S}_x}^{(l)},
\end{equation}
where $\mathcal{M}_x$ is the modality-aware attention mask.
The resulting features are mapped back to the corresponding Expert stream
using its output projection $W_{O,x}^{(l)}$.

The three Experts use a common attention-head geometry of 24 heads with a
head dimension of 128, while retaining modality-specific hidden dimensions
of 3072, 512, and 1024 for the Video, Motion, and Action streams,
respectively.

\subsubsection{Attention Connectivity and Motion-to-Action Ablation}

As described in Sec.~III.C, Motion and Action queries can attend to the
current visual condition but cannot access future visual tokens.
The Motion and Action streams can additionally exchange information through
the layer-wise mixed-attention mechanism.

To isolate the contribution of predictive motion to action generation, the
\textit{w/o Motion$\rightarrow$Action} variant removes Motion keys and
values from those accessible to Action queries. Specifically, the full model
uses
\begin{equation}
m\in\mathcal{S}_a,
\end{equation}
whereas the ablated variant uses
\begin{equation}
m\notin\mathcal{S}_a,
\end{equation}
equivalently masking the Motion-to-Action attention block as
\begin{equation}
\mathcal{M}_{A\leftarrow M}=-\infty.
\end{equation}

The reverse Action-to-Motion connection and all other attention paths remain
unchanged. The Motion Expert initialization, motion prediction objective,
target-robot demonstrations, optimization settings, and remaining model
components are identical to the full model.

\FloatBarrier

\subsection{Details of Experimental Setup}

\subsubsection{Details of Data Collection}

\begin{figure}[!t]
    \centering
    \includegraphics[width=0.9\columnwidth]{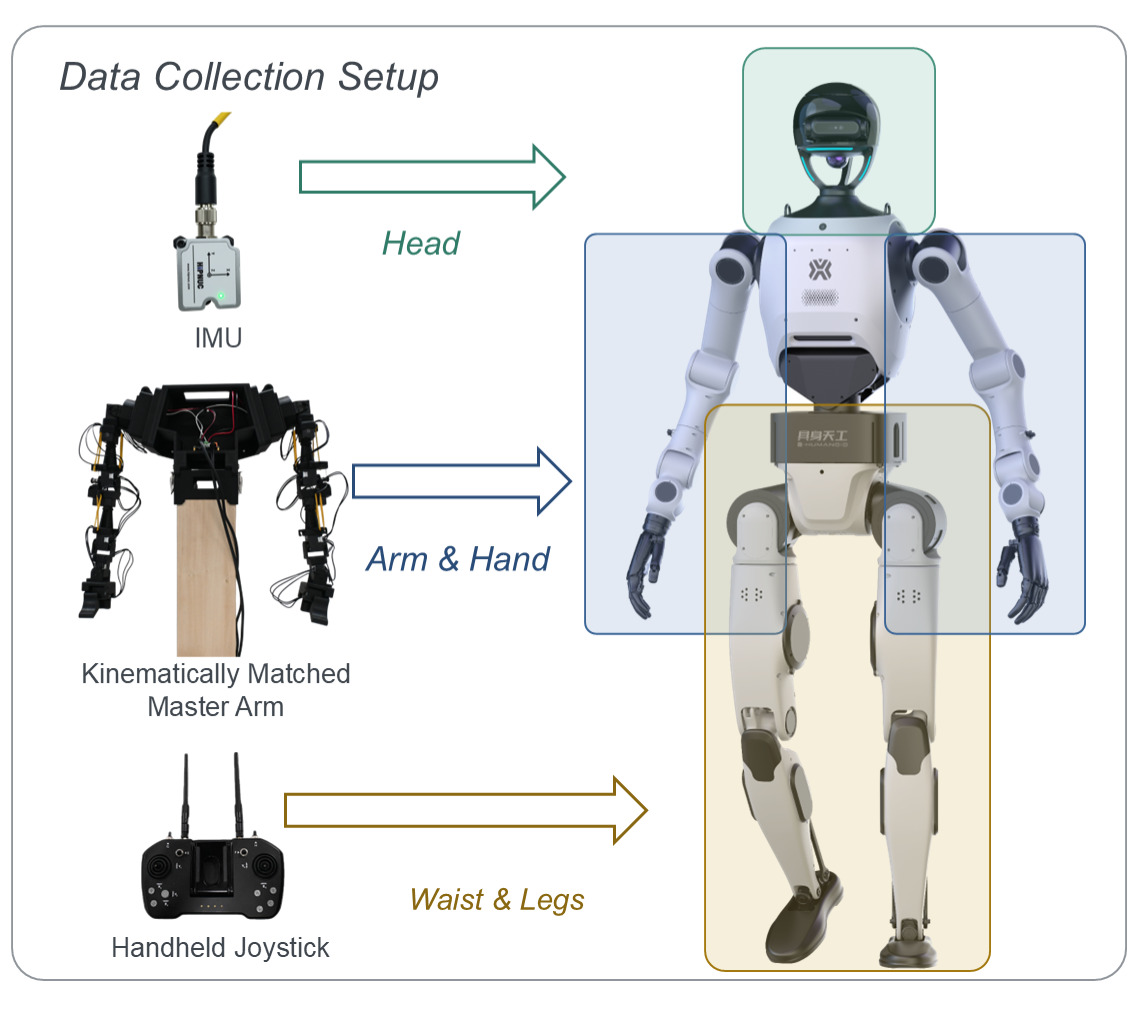}
    \vskip -0.15in
    \caption{\textbf{Data-collection setup of TianGong 3.0 platform.}
    An IMU controls head motion, kinematically matched master arms control
    the arms and end effectors, and a handheld joystick controls locomotion
    and waist motion.}
    \label{fig:data_collection}
\end{figure}

All target-robot demonstrations and real-world evaluations are conducted on
the TianGong 3.0 full-sized humanoid, equipped with an onboard egocentric RGB
camera for visual observations. Whole-body demonstrations are collected using
the teleoperation system shown in Fig.~\ref{fig:data_collection}.
A wearable IMU controls head motion, kinematically matched master arms
control the robot arms and end effectors, and a handheld joystick controls
locomotion and waist motion. This setup enables coordinated collection of
posture transitions, locomotion, and manipulation within each demonstration.

We collect 100 target-robot demonstrations per task and use the same
demonstration set for all compared methods.

\subsubsection{Real-Robot Evaluation Protocol}

Table~\ref{tab:appendix_tasks} summarizes the evaluation tasks and their
milestone-based scoring rules. Each completed milestone contributes the
specified number of points, and milestones are evaluated sequentially:
a later milestone receives credit only if all preceding milestones have been
completed.

\begin{table*}[!t]
\centering
\caption{Sequential milestone-based scoring protocol for real-robot evaluation.}
\label{tab:appendix_tasks}
\vskip -0.1in
\small
\begin{tabular}{p{0.18\textwidth} c p{0.55\textwidth} c}
\toprule
\textbf{Task}
& \textbf{\# Rollouts}
& \textbf{Sequential Milestones}
& \textbf{Max} \\
\midrule

Toy Pickup
& 20
& Grasp each doll: +5; place each doll into the target container: +5.
& 20 \\

Laundry Loading
& 20
& Grasp each garment: +2.5; load each garment: +2.5; close the lid: +5.
& 15 \\

Pillow Transfer
& 20
& Grasp pillow: +5; reach target location: +5; collision-free placement: +5.
& 15 \\

Kneeling Toy Storage
& 20
& Kneel: +2.5; grasp: +5; return to standing: +2.5; place into target box: +5.
& 15 \\

Toy Transfer
& 20
& Grasp: +5; transport to target: +5; place: +5.
& 15 \\

Box Transfer
& 20
& Lift: +5; transport to target: +5; place: +5.
& 15 \\

\bottomrule
\end{tabular}
\end{table*}

For each rollout, the points of all successfully completed milestones are
accumulated to obtain the rollout score $s_{t,i}$. To make scores comparable
across tasks with different maximum points, we normalize the accumulated
scores by the maximum achievable score:
\begin{equation}
\mathrm{Score}_t(\%)
=
\frac{\sum_{i=1}^{N_t}s_{t,i}}
{N_t s_t^{\max}}
\times 100,
\end{equation}
where $N_t$ is the number of rollouts for task $t$ and $s_t^{\max}$ is the
maximum score achievable in one rollout. All methods use the same task
definitions, initial-condition protocol, robot-side control stack, and
scoring procedure.

Toy Pickup requires repeated bending, grasping, and placement of two objects.
Laundry Loading combines deformable-object manipulation with
final interaction with the washing machine. Pillow Transfer requires
transporting a large object while maintaining collision-free whole-body
motion. Kneeling Toy Storage requires a kneeling-to-standing posture
transition during manipulation. Toy Transfer and Box Transfer require object
acquisition, transport, and final placement. Moderate object or container
variations are introduced in the relevant evaluation tasks.

\subsubsection{High-Level Action Interface}

The policy operates in a 34-dimensional whole-body action space comprising
14 arm-joint commands, 2 end-effector commands, 2 head commands,
6 posture-related commands, 3 waist commands, 6 locomotion-related
leg-velocity commands, and 1 mode/status command. The high-level policy does
not directly predict motor torques; its outputs are dispatched to the
corresponding TianGong 3.0 robot-side control interfaces.

\subsubsection{RL-Based Low-Level Whole-Body Controller}

Upper-body manipulation, head, and waist commands are executed through the
corresponding robot-side control interfaces, while locomotion and
posture-related commands are handled by dedicated robot-specific RL-based
low-level controllers. Walking behaviors use an AMP-style controller,
whereas kneeling and associated posture transitions use a BeyondMimic-style
motion-tracking controller.

All compared high-level policies share the same low-level control stack.
Therefore, differences in real-robot performance primarily reflect the
high-level policy rather than the underlying locomotion or balance
controllers.

\subsubsection{Implementation and Deployment Details}

Table~\ref{tab:appendix_training} summarizes the optimization settings for
Stage I and Stage II.
\begin{table}[!t]
\centering
\caption{Training configuration for Stage I and Stage II.}
\label{tab:appendix_training}
\vskip -0.1in
\small
\begin{tabular}{lcc}
\toprule
\textbf{Configuration} & \textbf{Stage I} & \textbf{Stage II} \\
\midrule
Optimizer & AdamW & AdamW \\
$\beta_1,\beta_2$ & $(0.9,0.95)$ & $(0.9,0.95)$ \\
Weight decay & 0.02 & 0.01 \\
Learning rate & $1\times10^{-5}$ & $1\times10^{-5}$ \\
LR schedule & Cosine restarts & Cosine \\
Warmup & $\sim$5\% & $\sim$5\% \\
Per-GPU batch size & 256 & 4 \\
Number of GPUs & 32 & 8 \\
Global batch size & 8192 & 32 \\
Training steps & 100K & 20K \\
History horizon & 16 & 16 \\
Future motion horizon & 32 & 32 \\
\bottomrule
\end{tabular}
\vspace{-0.4em}
\end{table}

During Stage I, the Motion Expert is trained on UniMotion-4K with precomputed
language embeddings from the frozen text encoder. The embodiment-specific
motion adapter $G_e$ is trained separately and remains frozen thereafter.

During Stage II post-training, the Motion Expert is initialized from Stage I
and jointly optimized with the Video and Action Experts. The video VAE and
embodiment-specific motion adapter remain frozen, while text embeddings are
precomputed and the text encoder is not loaded.

For target-robot motion conditioning, the 49-dimensional robot state is
reduced to the corresponding 29-dimensional joint state, mapped through the
frozen adapter $G_e$, and normalized to obtain the unified 63D
representation. Motion history is constructed exclusively from observed
robot states rather than predicted motion.

At deployment, a 30\,Hz proprioceptive buffer maintains the latest robot
states, with the most recent 16 forming $M_t^{\mathrm{hist}}$. At each policy
query, \method samples a 32-frame future-motion trajectory and a 32-step
action chunk using four Euler steps. The action chunk is executed at
30\,Hz, after which new visual and proprioceptive observations are acquired
and the policy replans. Predicted future motion is used only as a predictive
representation and is never fed back into the motion history or retargeted
into executable commands. Future visual latents are used only during
Stage II training and are omitted at deployment.

\subsubsection{Implementation Details of Baselines}

We compare \method with FastWAM~\cite{fastwam2026},
$\pi_{0.5}$~\cite{pi05_2025}, GR00T N1.7~\cite{grootn1_2025},
and $\tau_0$-WM~\cite{tau0wm2026}. FastWAM and $\tau_0$-WM serve as
representative world-action baselines, while $\pi_{0.5}$ and GR00T N1.7
provide representative VLA and humanoid foundation-model baselines.

All baselines are adapted to the same TianGong 3.0 policy-facing action
interface and trained using the same target-robot demonstration dataset.
We follow the corresponding official implementations and recommended
training recipes where applicable. During evaluation, all methods use the
same robot-side control stack, including the same low-level whole-body
controller, task configurations, and milestone-based scoring protocol.

\FloatBarrier

\subsection{Additional Experiments}

\subsubsection{Stage I Motion Prediction Capability}

We evaluate whether Stage I pretraining alone equips the Motion Expert with
robust predictive whole-body dynamics before any target-robot post-training.
Evaluation is performed on a held-out split of UniMotion-4K excluded from
Stage I training and entirely independent of the TianGong 3.0 demonstrations
used in Stage II.
\begin{table}[!t]
\centering
\caption{\textbf{Stage I future-motion prediction performance.}
Results are reported on the held-out UniMotion-4K split using MPJRE
($^\circ$); lower is better.}
\label{tab:stage1_motion_prediction}
\vspace{-0.3em}
\small
\begin{tabular}{lcc}
\toprule
\textbf{Method}
& \textbf{H16} $\downarrow$
& \textbf{H32} $\downarrow$ \\
\midrule
Static & 1.855 & 3.220 \\
Motion Expert, 1K h & 1.596 & 3.093 \\
Motion Expert, 2K h & 1.443 & 2.837 \\
Motion Expert, 4K+ h & \textbf{1.324} & \textbf{2.573} \\
\bottomrule
\end{tabular}
\vspace{-0.4em}
\end{table}
Given 16 historical motion frames, we evaluate prediction over the first
16 future frames (H16) and the full 32-frame horizon (H32). We report mean
per-joint rotation error (MPJRE, degrees), where lower is better.
As a non-predictive baseline, \textit{Static} repeats the last observed pose
throughout the prediction horizon.

As shown in Table~\ref{tab:stage1_motion_prediction}, all pretrained Motion
Experts outperform the static baseline, indicating predictive dynamics
beyond simple pose propagation. Performance also improves consistently with
pretraining scale. At 4K+ hours, MPJRE decreases from
1.855$^\circ$ to 1.324$^\circ$ at H16 and from
3.220$^\circ$ to 2.573$^\circ$ at H32, corresponding to reductions of
28.6\% and 20.1\%, respectively. These results are obtained before
target-robot post-training, showing that predictive whole-body dynamics are
acquired during Stage I.

\FloatBarrier